\documentclass[conference]{IEEEtran}

\usepackage[utf8]{inputenc}
\usepackage[T1]{fontenc}
\usepackage[russian,english]{babel}
\usepackage{parskip}

\usepackage{graphicx, float, subfigure, blindtext}

\newcommand\IEEEhyperrefsetup{
bookmarks=true,bookmarksnumbered=true,%
colorlinks=true,linkcolor={black},citecolor={black},urlcolor={black}%
}

\usepackage[\IEEEhyperrefsetup, pdftex]{hyperref}

\usepackage{mathtools}

\usepackage[nolist,nohyperlinks]{acronym}
\usepackage{cleveref}
\graphicspath{{images/}}
\usepackage[dvipsnames]{xcolor}

\acrodef{IC}[IC]{Integrated Circuit}
\newacro{svm}[SVM]{Support Vector Machine}
\author{
\IEEEauthorblockN{
    Alex Shonenkov
}
\IEEEauthorblockA{
    Sber AI, MIPT\\
    shonenkov@phystech.edu\\
    Kaggle: \href{http://kaggle.com/shonenkov}{@shonenkov}
}
\and
\IEEEauthorblockN{
    Daria~Bakshandaeva
}
\IEEEauthorblockA{
    Sber AI\\
    DDBakshandaeva@sberbank.ru
}
\and
\IEEEauthorblockN{
    Denis~Dimitrov
}
\IEEEauthorblockA{
    Sber AI, Lomonosov MSU\\
    Dimitrov.D.V@sberbank.ru
}
\and
\IEEEauthorblockN{
    Aleksandr~Nikolich
}
\IEEEauthorblockA{
    Sber AI, MIREA\\
    ADrNikolich@sberbank.ru
}
}

\title{Emojich – zero-shot emoji generation using Russian language: a technical report}
\begin{document}
\maketitle
\begin{figure}[t!]
    \centering
    \includegraphics[width=0.48\textwidth]{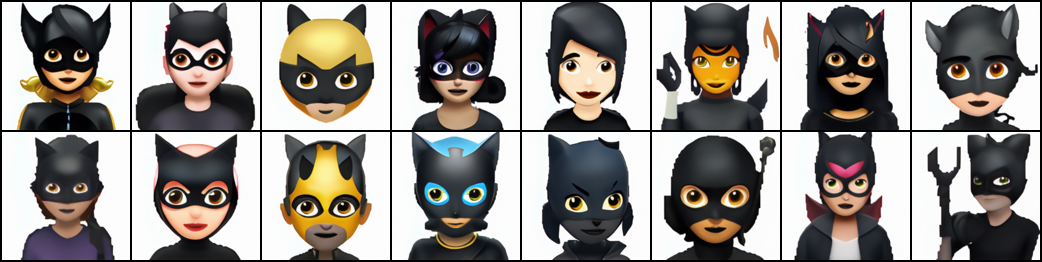}
    \caption{
        Emojis generated by the text condition \selectlanguage{russian}``женщина-кошка''\selectlanguage{english} (``catwoman'')
    }
    \label{fig:my_label}
\end{figure}

\begin{abstract}
This technical report presents a text-to-image neural network ``Emojich'' that generates emojis using captions in Russian language as a condition. We aim to keep the generalization ability of a pretrained big model ruDALL-E Malevich (XL) 1.3B parameters at the fine-tuning stage, while giving special style to the images generated. Here are presented some engineering methods, code realization, all hyper-parameters for reproducing results and a Telegram bot where everyone can create their own customized sets of stickers. Also, some newly generated emojis obtained by "Emojich" model are demonstrated.
\end{abstract}
\begin{center}
\begin{IEEEkeywords}
text-to-image, emoji, deep learning, ruDALL-E
\end{IEEEkeywords}
\end{center}

\section{Introduction}
\label{section:intro}
The transformer models which are normally pretrained on a large amount of data, have proved their ability to cope successfully with the various downstream tasks or adapt to new data related to the specific domain – and this is achieved through fine-tuning. Examples of this approach are abundant in the fields of Natural Language Processing \cite{DBLP:conf/coling/GriesshaberMV20}, \cite{DBLP:journals/corr/abs-2106-10199}, \cite{DBLP:conf/iclr/MosbachAK21} and Computer Vision \cite{DBLP:journals/corr/abs-2109-13925}, \cite{DBLP:journals/corr/abs-2110-05270}. However, multimodal models are currently at the cutting edge of ML research, with text-to-image pretrained transformer models \cite{DBLP:conf/icml/RameshPGGVRCS21}, \cite{DBLP:journals/corr/abs-2105-13290}, \cite{rudalle_github} being one of the brightest examples in the field. As for improving the fine-tuning process of such models, there is still considerable latitude for the experiments.

It is rather time-consuming and challenging procedure\footnote{\url{https://unicode.org/emoji/proposals.html}} – to prepare and submit a proposal for a new emoji to the Unicode Emoji Subcommittee; most importantly, the success is not guaranteed. Emojis – with its expressiveness, universality and ease of use – has infiltrated into our everyday digital communication. It would be a great option, to generate unique and customize emojis on the fly, especially since some messengers (such as Telegram) allow users to create their own sets of icons. In this regard, recently introduced text-to-image models which compute joint distribution over texts and images and generate the latter by the former, seem very appealing. Similar efforts have already been undertaken in \cite{DBLP:journals/corr/abs-1712-04421}, however, some limitations can be observed there: the proposed approach is based on DC-GANs and word2vec embeddings, whereas the transformer architecture of the recent models provides high-quality representations which depend on the context and capture the meaning much better (for instance, in \cite{DBLP:journals/corr/abs-1712-04421} the samples generated on more than 1 word embedding or some new unexpected text prompts were very noisy), and allows for more deep interaction between modalities; the data used for training is rather poor and restricted to 82 emotive faces; the resulted model cannot synthesize new emojis (and the best prospect for it, as seen by the authors, is to generate icons which are the combination of the existing ones). In view of the above, we would like to fine-tune ruDALL-E model in order to test its ability to generate emojis based on the textual descriptions – just like an artist doing commissions to delight the customers.

\section{Dataset}
\label{section:dataset}
The proposed Emoji dataset is collected by web scrapping: emoji icons with the corresponding text captions in Russian are retrieved. The emoji pictures that are naturally in RGBA format with the last channel (alpha) being an opacity channel, are converted to RGB: pixels with opacity value less than 128 are assigned white color code. Small emoji pictures are scaled to 256 px using super-resolution approach Real-ESRGAN \cite{wang2021realesrgan}, that has been trained by Sber AI with ``$\times$4'' scaling \cite{realesrgan_sberai}. 

The full version of the dataset is available on Kaggle\footnote{\url{https://www.kaggle.com/shonenkov/russian-emoji}}. The dataset contains 2749 images and 1611 unique texts. Such mismatch is due to the following reasons: there are sets, within which emojis differ only in color (for instance, images of hands with different skin tones); some elements are homonyms in Russian (for example, the descriptions for the images of a medieval castle and a padlock are identical – \selectlanguage{russian}{``замок''}). \selectlanguage{english}The length of text descriptions varies from 1 to 7 words (about 67\% of all samples are one- or two-word descriptions).

\begin{figure}[t!]
    \centering
    \includegraphics[width=0.3\textwidth, height=5cm]{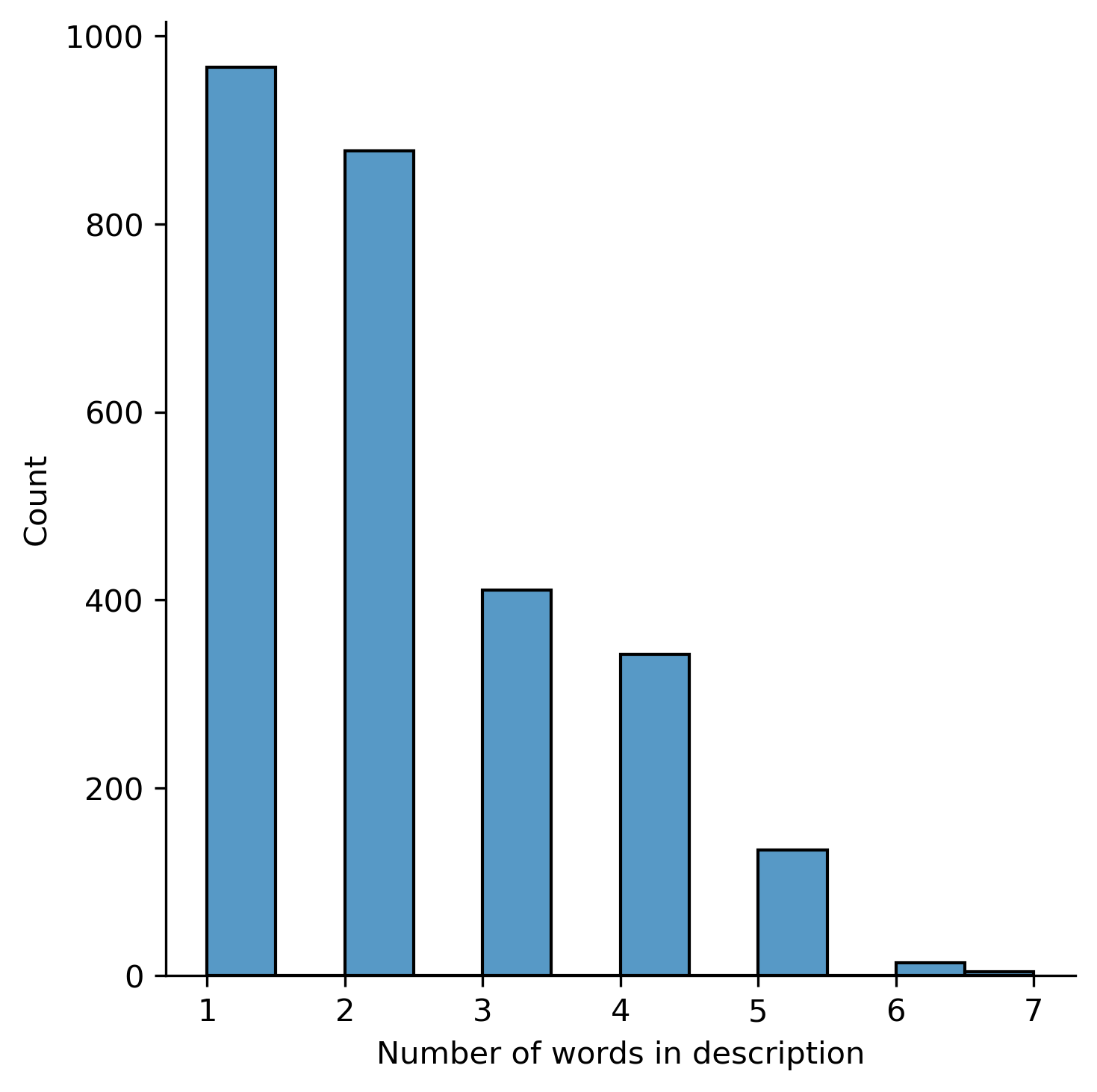}
    \caption{
        Distribution of word count values per description in the Emoji dataset
    }
\end{figure}

\section{Fine-Tuning}
\label{section:method}

The main goal of fine-tuning is to keep the ``knowledge of the world'' of ruDALL-E Malevich (XL) model while adjusting the style of generated images to emoji's. ruDALL-E Malevich is a big multimodal pretrained transformer, which learns the correlations between images and texts \cite{rudalle_github}. The freezing of the feedforward and self-attention layers in a pretrained transformer has demonstrated high performance in a multimodal setup \cite{lu2021pretrained}, \cite{bakshandaeva2021heads}. Thus, by freezing our model we will definitely avoid catastrophic forgetting, increase the efficiency of the process and reduce GPU memory usage.

The proposed Emoji dataset has limited variations of captions (see chapter~\ref{section:dataset} for more details), therefore the model has all chances to be overfitted on the text modality and lost generalization. To deal with this issue, the coefficient of the weighted cross-entropy loss is increased to $10^3$ for image representations (the codebook vectors). No data augmentations are used.

8-bit Adam \cite{dettmers20218bit} is used for optimization. This realization reduces the amount of GPU memory needed for gradient statistics and provides more stable training with fp16 precision. One cycle learning rate is chosen as a scheduler with the following parameters: start lr 4e-7, max lr 1e-5, final lr 2e-8, warmup 0.1, epochs 40, batch size 2, gradient clipping 1.0. Training time is 5h 30m using 1xA100.

The source code used for fine-tuning is available on Kaggle\footnote{\url{https://www.kaggle.com/shonenkov/emojich-rudall-e}}.

\begin{figure}[H]
    \centering
    \subfigure[Training loss]{
    \label{fig:train_loss}
    \includegraphics[width=.46\columnwidth]{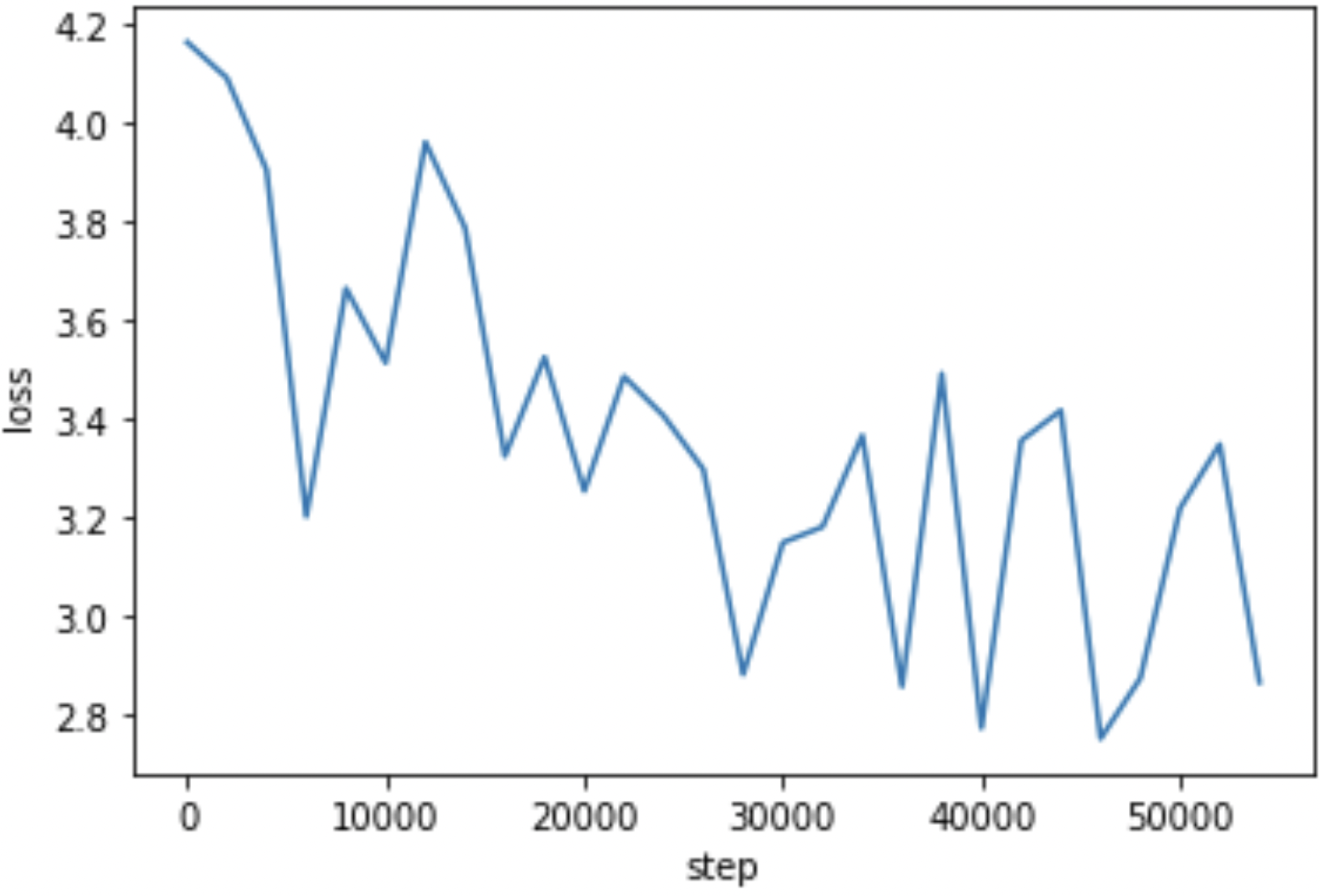}}
    \subfigure[Learning rate]{
    \label{fig:lr}
    \includegraphics[width=.46\columnwidth]{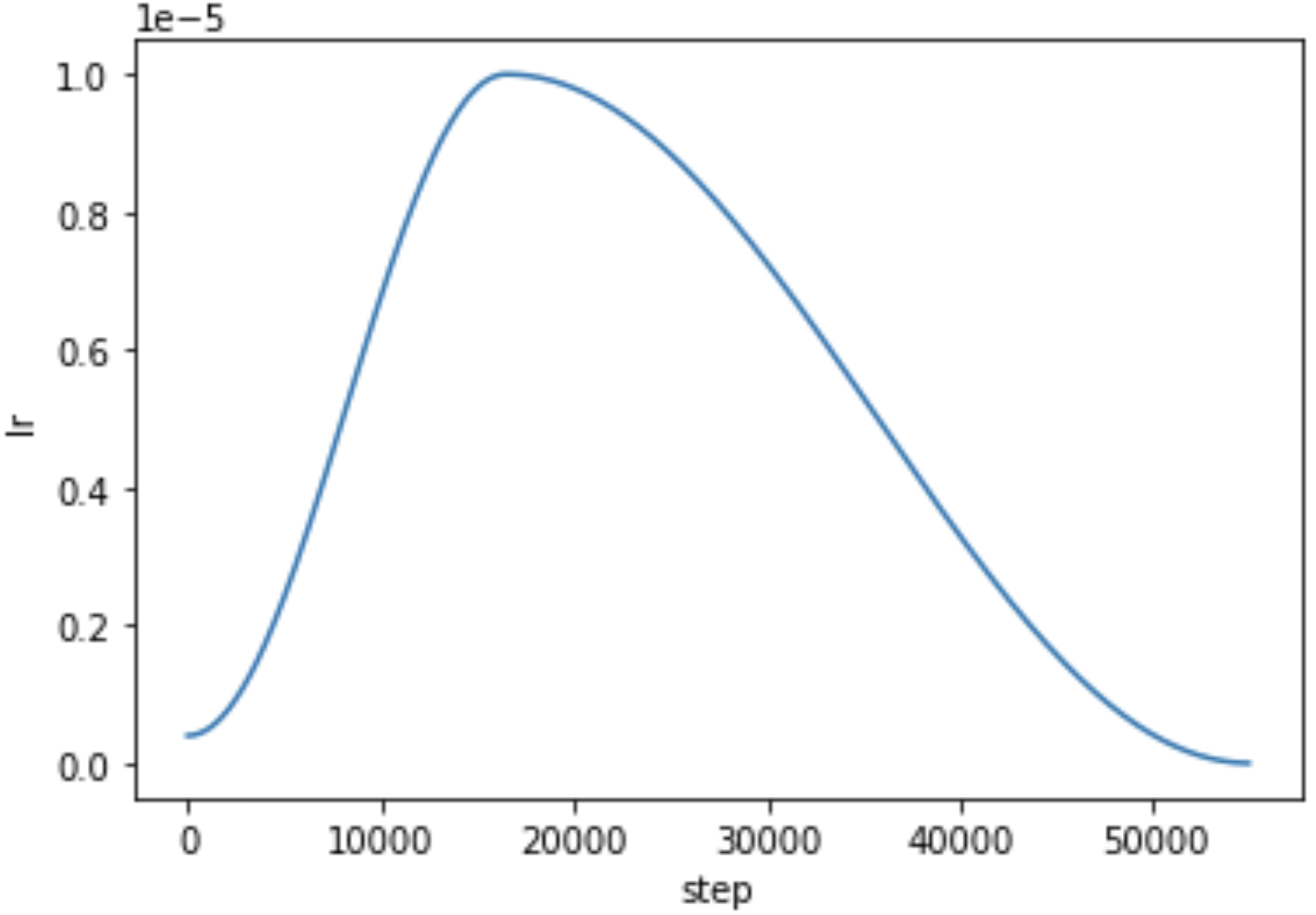}}
    \caption{Some measurements during fine-tuning.}
\end{figure}

\section{Evaluation}
\label{section:eval}
A regular evaluation is an essential part of the fine-tuning process (this holds true, without a doubt, for pretraining as well) as it indicates when the fine-tuning should be stopped. In case of image generation, the most popular metrics is FID \cite{DBLP:conf/nips/HeuselRUNH17} which is used to evaluate the quality of Generative Adversarial Networks (GANs) performance. After obtaining the representations for both sets of artificial and real images via the Inception Net, two multivariate Gaussians fitted to these sets of feature embeddings are compared by computing Fréchet distance between them. It is rightly noted \cite{DBLP:journals/corr/abs-2105-13290} that this metrics is suited for evaluation of general-domain unconditional generation from simple distributions, thereby it is not the best option in our case. 

In \cite{DBLP:journals/corr/abs-2105-13290} another metric is proposed – Caption Loss. It involves fine-tuning the model for the inverse task – image captioning – and further self-reranking using cross-entropy values for the text tokens. This metric is worth considering and will be used in further experiments. 

\begin{figure}[t!]
    \centering
    \includegraphics[width=0.48\textwidth]{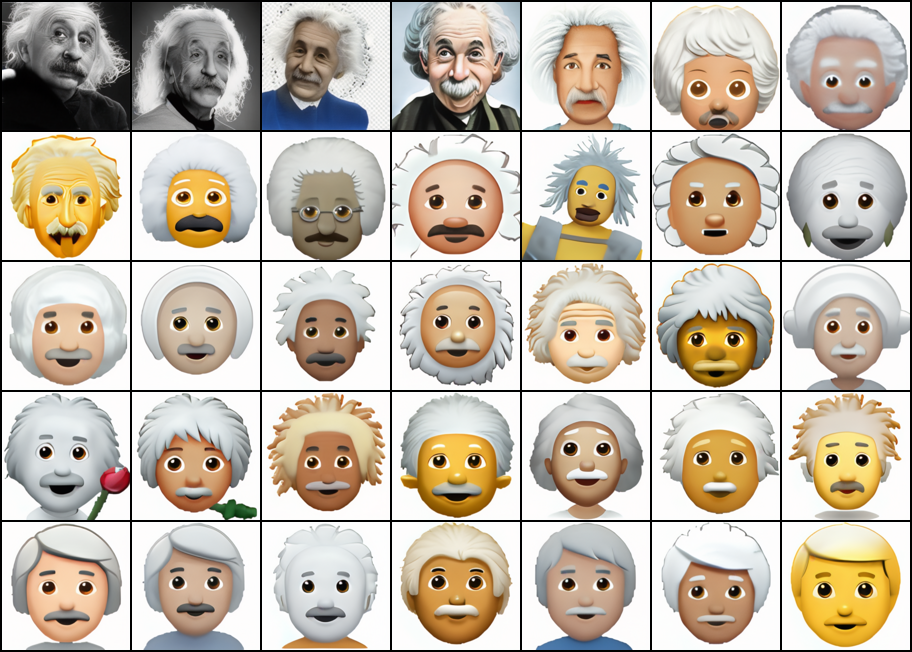}
    \caption{
        From the top left corner to the lower right corner: images generated by the model after every 2 thous. iterations of fine-tuning using the text prompt \selectlanguage{russian}``Эйнштейн улыбается''\selectlanguage{english} (``Einstein is smiling'')
    }
    \label{fig:albert}
\end{figure}

In the current experiments, human evaluation is used: every 6 thousandths iterations of fine-tuning, 64 images generated by the model are assessed using such criteria as the degree of generality and abstraction, conformity to the emoji aesthetics and overall quality of the image (for instance, smoothness and absence of glitches/artefacts). Figure~\ref{fig:albert} demonstrates generated image samples obtained by the model after every 2 thous. iterations of fine-tuning: starting with the original ruDALL-E Malevich (XL) (in the upper left corner) and finishing with the model that has been fine-tuning for 68 thous. iterations in total (in the lower right corner). It can be observed that initially images are realistic photos that do not correspond to the emoji style; later they are modified and resemble drawings; after about 12 thous. iterations images start to look like emojis more and more (although there could be some features or artefacts that are out of style – a pose, for example) while maintaining the recognizable traits of Einstein; after about 56 thous. iterations, however, the model is consistently loosing the ``knowledge of the world'': the images are becoming uniform and impersonal – Einstein is transforming to an unknown man with a moustache, the standard emojis are starting to prevail in the generation results.

\section{Emoji generation}
\label{section:generation}

The emoji generation starts with a text prompt that describes the desired emoji content. When the tokenized text is fed to Emojich, the model generates the remaining codebook vectors auto-regressively. Every codebook vector is selected item-by-item from a predicted multinomial probability distribution over the image latents using nucleus top-p and top-k sampling with a temperature \cite{DBLP:journals/corr/abs-1904-09751} as a decoding strategy. The image is rendered from the generated sequence of latent vectors by the decoder of the dVAE, which is pretrained VQ-GAN \cite{esser2020taming} with Gumbel Softmax Relaxation \cite{kusner2016gans}. Also, the decoder is modified with the inverse discrete wavelet transform (IDWT) used in the neural network MobileStyleGAN \cite{DBLP:journals/corr/abs-2104-04767}; it allows restoring 512$\times$512 (instead of 256$\times$256) images almost without loss of quality. The authors would like to thank @bes-dev for his help with the implementation of this modification\footnote{\url{https://github.com/bes-dev/vqvae_dwt_distiller.pytorch}}.

\section{Emoji segmentation}
\label{section:segmentation}
In order to use images generated by Emojich as emoji icons or stickers in messengers, they should be converted to RGBA format. However, the problem is not as straightforward as assigning alpha channel's value to 0 to all white colored pixels: some inner parts of the image can be white and should not be made transparent. Thus, the problem of image segmentation arises. 

U-Net \cite{ronneberger2015unet} with EfficientNetB7 as an encoder is used as a binary semantic segmentation model (with 0 meaning complete transparency and 1 – complete opacity). The training RGBA dataset is obtained using pseudo-labeling method: at first, U-Net model is trained on the initial dataset of emojis in RGBA format (2749 pictures); then predictions on the emojis generated by Emojich are made by the model – the predicted segmented images, which confidence value is not less than 0.99, are added to the training data; the process repeats until the training reaches the plateau. Some nontrivial cases are segmented manually using classical contour-based approach \footnote{\url{https://docs.opencv.org/3.4/d4/d73/tutorial_py_contours_begin.html}} and further are added to the training set as well. The size of the final train dataset is 246089 images. All images of the validation set (143 icons) are segmented manually. 

The resulting pipeline involves two possible regimes: the generated image is segmented by the trained U-Net model; if the confidence value is less than the threshold (0.99), classical contour-based approach is used.

\section{Results}
\label{section:results}

In the appendix, some emoji pictures created through the proposed methods are demonstrated. All examples are generated automatically (without manual cherry-picking) with the following hyper-parameters: seed 42, batch size 16, top-k 2048, top-p 0.995, temperature 1.0, GPU A100. To obtain emojis of better quality we advise to use more attempts ($\sim512$) and select the best one manually. Remember, for the great art makers creating just one masterpiece is enough to become ``great''.
 
 Moreover, we propose a segmentation procedure based on U-Net to crop the background of the generated images of the emojis. It is necessary to create looking good stickers. Thus everyone can generate their own customized sets of stickers using this Telegram bot: https://t.me/rudalle\_emojich\_bot.

\section*{Acknowledgment}
The authors would like to thank Sber AI, SberDevices and SberCloud for granting the GPU-resources for the experiments. 

\bibliographystyle{IEEEtran}
\bibliography{references}

\cleardoublepage


\onecolumn

\begin{figure}[H]
    \centering
    
    \subfigure[\selectlanguage{russian}"зима в горах"\selectlanguage{english} ("a winter in the mountains")]{
    \includegraphics[width=1.0\columnwidth]{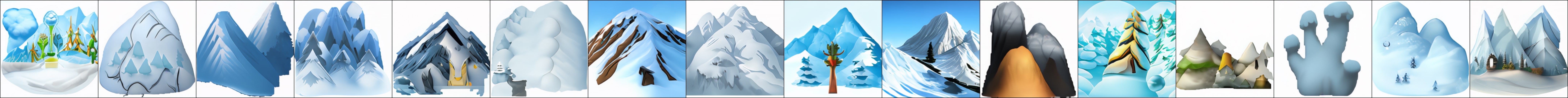}}
    
    \subfigure[\selectlanguage{russian}"флаг зомби апокалипсиса"\selectlanguage{english} ("a zombie apocalypse flag")]{
    \includegraphics[width=1.0\columnwidth]{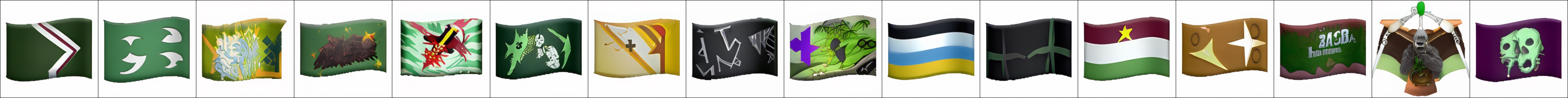}}
    
    \subfigure[\selectlanguage{russian}"флаг Cбербанка"\selectlanguage{english} ("Sberbank flag")]{
    \includegraphics[width=1.0\columnwidth]{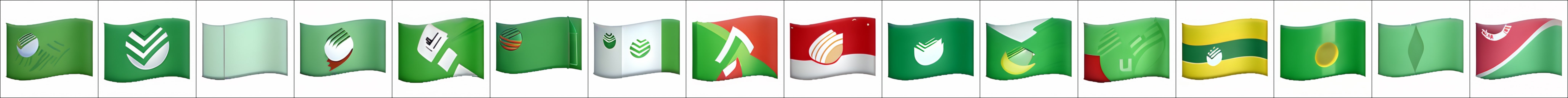}}
    
    \subfigure[\selectlanguage{russian}"храм Василия Блаженного"\selectlanguage{english} ("St. Basil's Cathedral")]{
    \includegraphics[width=1.0\columnwidth]{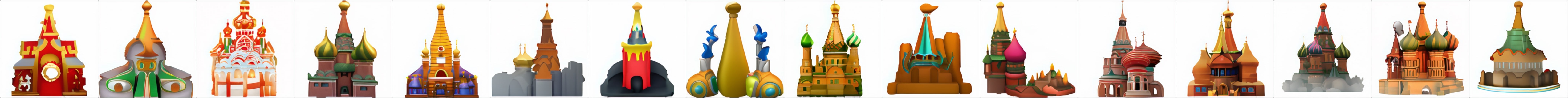}}    
    
    \subfigure[\selectlanguage{russian}"вишневая девятка"\selectlanguage{english} ("cherry-colored Lada 2109")]{
    \includegraphics[width=1.0\columnwidth]{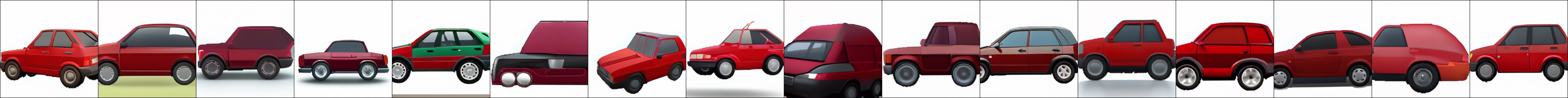}}    
    
    \subfigure[\selectlanguage{russian}"Дональд Трамп из лего"\selectlanguage{english} ("Donald Trump made of LEGO")]{
    \includegraphics[width=1.0\columnwidth]{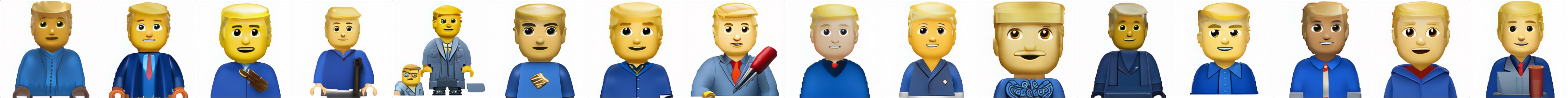}}
    
    \subfigure[\selectlanguage{russian}"человек ест яблоко"\selectlanguage{english} ("a human eats an apple")]{
    \includegraphics[width=1.0\columnwidth]{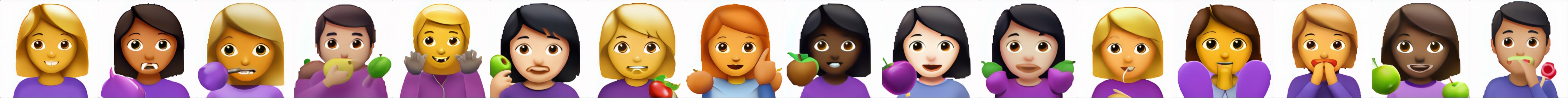}}

    \subfigure[\selectlanguage{russian}"ежик в голубой шапке"\selectlanguage{english} ("a hedgehog in a blue hat")]{
    \includegraphics[width=1.0\columnwidth]{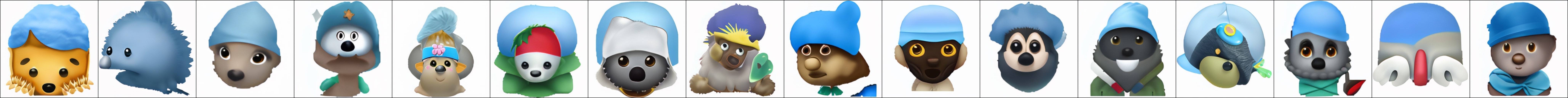}}

    \subfigure[\selectlanguage{russian}"волк в овечьей шкуре"\selectlanguage{english} ("a wolf in sheep's clothing")]{
    \includegraphics[width=1.0\columnwidth]{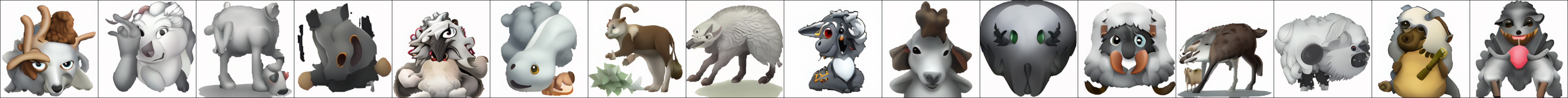}}
    
    \subfigure[\selectlanguage{russian}"кролик синего цвета"\selectlanguage{english} ("a blue rabbit")]{
    \includegraphics[width=1.0\columnwidth]{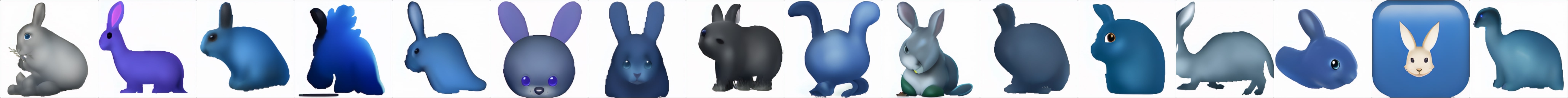}}
    
    \subfigure[\selectlanguage{russian}"розовая альпака улыбается"\selectlanguage{english} ("pink alpaca is smiling")]{
    \includegraphics[width=1.0\columnwidth]{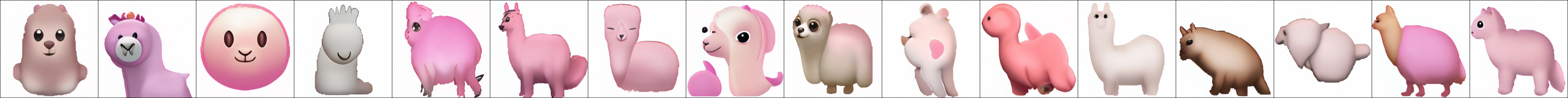}}
    
    \subfigure[\selectlanguage{russian}"арфа в форме улитки"\selectlanguage{english} ("a snail-shaped harp")]{
    \includegraphics[width=1.0\columnwidth]{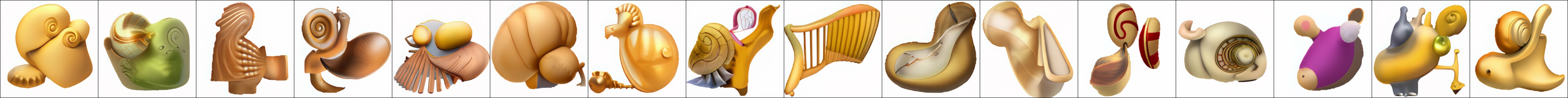}}
    
\end{figure}

\end{document}